\documentclass[10pt,twocolumn,letterpaper]{article}

\usepackage{cvpr}
\usepackage{times}
\usepackage{epsfig}
\usepackage{graphicx}
\usepackage{amsmath}
\usepackage{amssymb}
\usepackage{algorithm}
\usepackage{booktabs}
\usepackage{multirow}
\usepackage{multicol}
\usepackage{dsfont}

\usepackage[pagebackref=true,breaklinks=true,letterpaper=true,colorlinks,bookmarks=false]{hyperref}

\cvprfinalcopy % *** Uncomment this line for the final submission

\begin{document}

%%%%%%%%% TITLE
% \title{Adaptive Threshold for Better Performance of the Recognition and Re-identification Models}
\title{Adaptive Threshold for Online Object Recognition and Re-identification Tasks}

\author{Bharat Bohara\\
deHumlaTech AI Research \\
{\tt\small dehumlatech@gmail.com}
}

\maketitle
%\thispagestyle{empty}

%%%%%%%%% ABSTRACT
\begin{abstract}
    Choosing a decision threshold is one of the challenging job in any classification tasks. How much the model is accurate, if the deciding boundary is not picked up carefully, its entire performance would go in vain. On the other hand, for imbalance classification where one of the classes is dominant over another, relying on the conventional method of choosing threshold would result in poor performance. Even if the threshold or decision boundary is properly chosen based on machine learning strategies like SVM and decision tree, it will fail at some point for dynamically varying databases and in case of identity-features that are more or less similar, like in face recognition and person re-identification models. Hence, with the need for adaptability of the decision threshold selection for imbalanced classification and incremental database size, an online optimization-based statistical feature learning adaptive technique is developed and tested on the LFW dataset and self-prepared athletes dataset. This method of adopting adaptive threshold resulted in $12-45\%$ improvement in the model accuracy compared to the fixed threshold $\{0.3, 0.5,0.7\}$ that are usually taken via the hit-and-trial method in any classification and identification tasks. Source code for the complete algorithm is available at: \href{https://github.com/Varat7v2/adaptive-threshold.git}{https://github.com/Varat7v2/adaptive-threshold.git}
\end{abstract}


%%%%%%%%% BODY TEXT
\section{Introduction}

Most of the research is mainly focused on revising the performance of classification architectures, optimization algorithms, and loss functions, but hardly done any work on the improvement over adaptability of the decision threshold that separates the boundary-line between the classes. Although there are myriad state-of-the-art models to extract discriminative features, it is still a common trend to choose a classification threshold by the hit-and-trial method. As per the author's best practice, for a handful of feature vectors in the database, it works pretty well, but it becomes ineffective with the increasing database's size. If the chosen threshold is futile, how much accurate the model is, its state-of-the-art performance would go in vain. On the contrary, if the threshold is chosen wisely and is updated iteratively as soon as a new feature vector is registered in the database, even less accurate model will do a better job. In this project, an optimization-based statistical feature learning algorithm has been developed to boost-up the performance of any recognition and re-identification models through decision boundary.

A decision threshold is a numerical value that dichotomizes different classes. Different thresholds yield a different number of true/false positives and true/false negatives, and consequently different precision, recall, and f1-score for a given dataset. In this project, a decision threshold is selected to maximize the f1-score (that is, the harmonic average of precision and recall). For instance in cancer diagnostic problem, it acts as a specified cut-off for an observation to be classified as either 0 (no cancer) or 1 (has cancer). Choosing an optimal threshold value is a challenging task as it is case-specific, i.e., different for different objectives and datasets. In the context of verification and identification tasks, as the identity-database like in face recognition and person re-identification gets updated with a newer identity raises the need for tuning the threshold with database update. This threshold works not only as the deciding factor for verification and identification but also as the gatekeeper to update the identities in database.

\section{Related work}
There is not much work done yet in selecting the classifier threshold adaptively; however, Chou \textit{et. al} (2018) \cite{Chou} has developed an online face registration mechanism with a distinctive threshold per face. With this mechanism, they achieved 22\% accuracy improvement on the LFW dataset as compared to the common method of choosing the threshold - fixed value throughout the inference time. In brief, they have proposed an adaptive thresholding technique that assigns a specific threshold per registered face in the database – that gets adopted accordingly with the new entries in the database. However, this method is limited only to the evaluation task but not favorable to the real-time inference. 

Receiver operating characteristics (ROC) curve plots the true-positive rate against the false-positive rate at particular thresholds. A diagonal line indicates that the model predicts all the cases as the majority class delineating its poor performance over the test-dataset; above the curve indicates a threshold with higher model accuracy and below it with poor accuracy. ROC curve helps to understand the trade-off in the true-positive rate and false-positive rate for different threshold values. The area under the ROC curve is known as the area under the curve (AUC) that depicts the model's overall performance.

Zou \textit{et al.} (2016) \cite{Zou2016}, have analyzed drawbacks of using ROC-curve as the sole measure of selecting the threshold for an imbalanced classification. A novel framework -  sampling-based threshold auto-tuning method, for finding the best classification threshold is proposed yielding $20.63\%$ improvement over the conventional method of choosing a default threshold of 0.5. Though ROC curve and AUC values reflect ranking power of positive prediction probability, Zou \textit{et al.} claim that classifier performance including precision, recall, and f1-score might not be perfect even-though AUC value exceed 0.9. They have employed f1-score and AUC for Liao’s protein remote homology detection such that the classification threshold is tuned for the best f1-score. Similarly, Lipton \textit{et al.} (2014) \cite{Lipton2014} has proposed an optimal classification threshold selection through maximization of f1-score value for binary and multi-label classification problem - claiming improvement in predicting 26,853 labels of Medline documents compared to the traditional method.

Al Hartmann \cite{hartmann2009} has filed a patent on an adaptive threshold for detecting spam email messages based on ratios between clean and spam emails received at previous time-periods and misclassification ratio cost. Bauer \textit{et al.} (2015) \cite{Bauer2015} has applied a Bayesian model of neurofeedback and reinforcement learning for evaluating the impact of adaptive classification threshold on optimizing restorative brain-computer interfaces (BCI). It is claimed that a threshold adaptation is superior to any-of-the fixed threshold throughout the experimental results.

\section{Methodology}

For statistical feature learning of a probability density function, different metrics like cosine distance, cosine similarity, and euclidean distance as mentioned in equations [\ref{eqn:cosine_similarity},\ref{eqn:cosine_distance},\ref{eqn:euclidean_distance}] are used. Nonetheless similarity metric seems to fit best to our objective problem. Cosine similarity is a metric for measuring the distance between two vectors of an inner product space. In other words, it is the cosine of the angle between the two vectors projected in a multi-dimensional space, also commonly known as the inner dot product of those vectors - both normalized to have a unity length. The vectors' similarity is between 0 to 1; 0 means entirely dissimilar, and 1 means completely similar. So higher the value, higher will be the similarity between the vectors. Similarly, euclidean distance is a metric to measure distance or length between the two points in a euclidean space. The significance of all these metrics, especially in the machine learning arena, is to compute how similar or dissimilar the feature vectors are. In this project, since there are multiple feature vectors for the same identity (i.e., face or person), vectorial similarities and distances are organized into auto and cross-category - inspired from the auto/cross-correlation terminologies.

\begin{figure}[h]
\centering
\includegraphics[width = 0.48\textwidth]{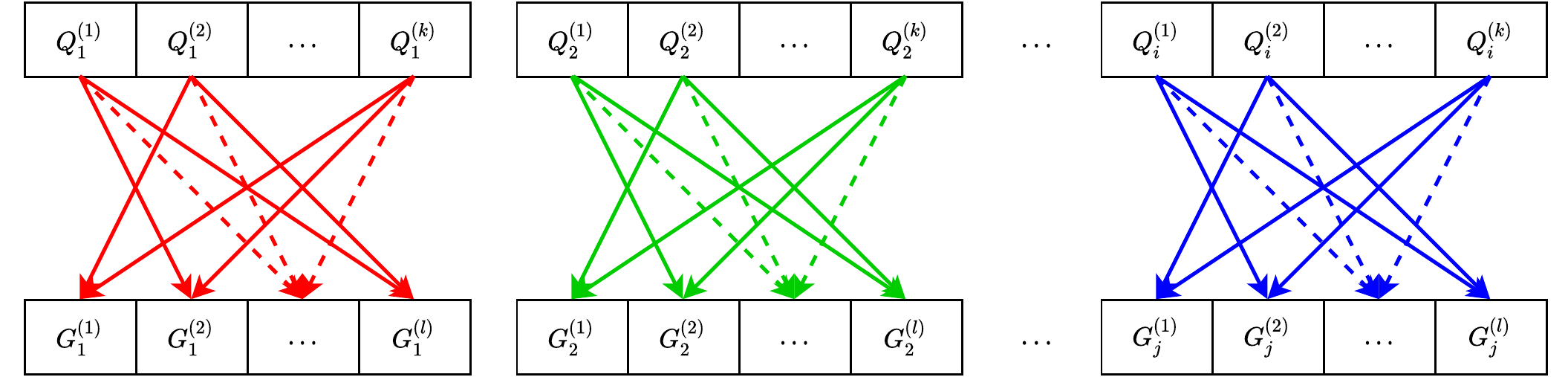}
\caption[Auto similarities pairing]{Auto similarities pairing}
\label{fig:auto_similarity_pairing}
\end{figure}

\begin{figure}[h]
\centering
\includegraphics[width = 0.48\textwidth]{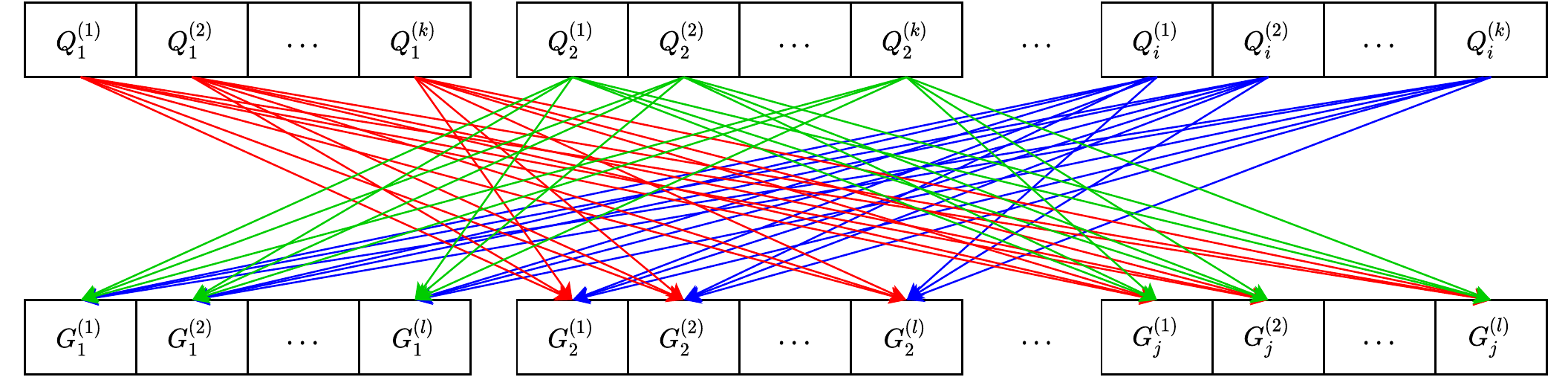}
\caption[Cross similarities pairing]{Cross similarities pairing}
\label{fig:cross_similarity_pairing}
\end{figure}

\begin{figure}[h]
\centering
\includegraphics[width = 0.48\textwidth]{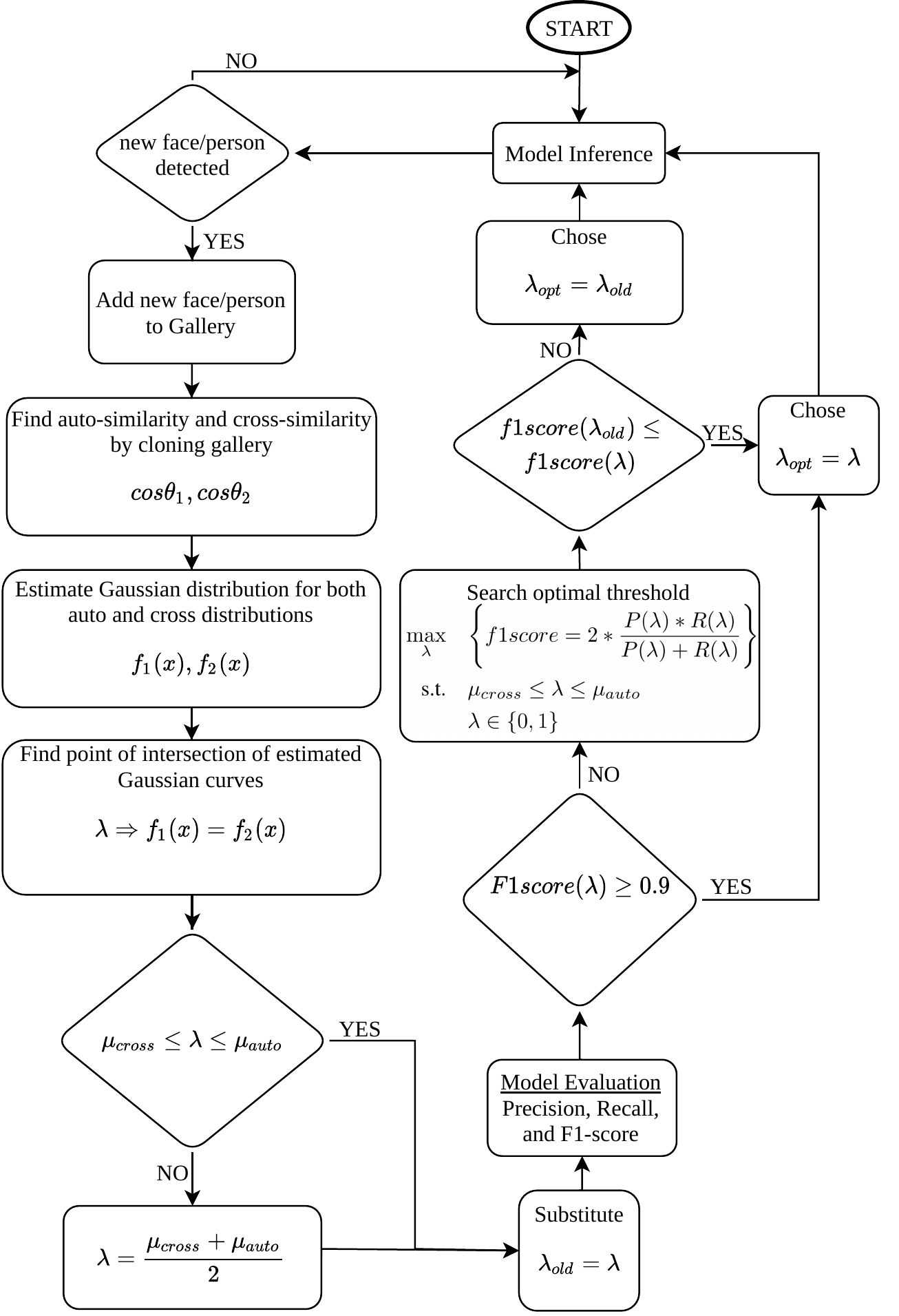}
\caption[Flow-diagram of online optimization based adaptive threshold search method]{Flow-diagram of online optimization based adaptive threshold search method}
\label{fig:flow_diagram_adaptive_threshold}
\end{figure}
Similarities and distances between the feature vectors of the same (here denoted as auto) and different (here denoted as a cross) identities (here identities denote faces, persons, or any other subject of interest) are computed simultaneously. Figures [\ref{fig:auto_similarity_pairing},\ref{fig:cross_similarity_pairing}] show all the possible pairing of the identities in order to compute auto and cross distance/similarity between the query, and database identities. While finding an adaptive threshold, query embeddings ($Q_{\imath}$) are cloned from gallery embeddings ($G_{\jmath}$) to find the similarity among same (auto-similarity) and different identities (cross-similarity). Hence during the query mode, $G_{\imath}$ and $Q_{\jmath}$ will be different images or embeddings, whereas during optimal-threshold search mode, query embeddings ($Q_{\imath}$) will be an exact copy of database embeddings ($G_{\jmath}$). Different statistical features like minima, maxima, mean, standard deviation, variance, and much more information can be drawn from auto and cross distance/similarity distributions. From such statistical information, an adaptive threshold will be adjusted once a new identity is added or deleted from the database. Since calculating statistical measures along with euclidean distance or cosine similarity between the feature vectors from a large pool of distributions is computationally expensive, an adaptive threshold adjustment can be done periodically once the number of newer identity registration exceeds a certain specified number.

\begin{equation} \label{eqn:cosine_similarity}
\begin{split}
similarity (\cos\theta) & = \frac{\textbf{X}\cdot\textbf{Y}}{\left\|\textbf{X}\right\|\left\|\textbf{Y}\right\|} \\
 & = \frac{\sum_{i=1}^{n}X_i Y_i}{\sqrt{\sum_{i=1}^{n} X_i^2} \sqrt{\sum_{i=1}^{n} Y_i^2}}
\end{split}
\end{equation}

% \begin{equation}
%     cosine-similarity (cos\theta) = \frac{\textbf{X}\cdot\textbf{Y}}{\left\|\textbf{X}\right\|\left\|\textbf{Y}\right\|} = \frac{\sum_{i=1}^{n}X_i Y_i}{\sqrt{\sum_{i=1}^{n} X_i^2} \sqrt{\sum_{i=1}^{n} Y_i^2}}
% \end{equation}

\begin{equation} \label{eqn:cosine_distance}
    D_{cosine} = 1 - \cos\theta
\end{equation}

\begin{equation} \label{eqn:euclidean_distance}
    D_{euclidean} (X,Y) = \sqrt{\sum_{i=1}^{n} \left(X_i - Y_i\right)^2}
\end{equation}

Referring to figure \ref{fig:flow_diagram_adaptive_threshold}, once auto and cross similarity distribution is obtained, their probability of occurrences is depicted in a histogram. With a mean and a standard deviation of auto and cross probability distributions, a Gaussian function is estimated for each of them, as shown in equation \ref{eqn:gaussian_functions}. The main reason behind estimating auto and cross Gaussian distribution function is to choose a value between their means such that none of the distribution is dominated or biased while choosing the threshold. We can either choose an average of auto and cross means directly without estimating their normal distribution, but doing so might unnecessarily shift our threshold towards a curve with a higher peak. In order to find the point of intersection of the estimated auto and cross Gaussian distribution functions, we need to equate and solve them, as shown in quadratic equation \ref{eqn:quadratic_equation}.

\begin{equation} \label{eqn:gaussian_functions}
  \begin{aligned}
  f_1(x) = \frac{1}{\sigma_{1} \sqrt{2\pi}} e^{-\frac{1}{2} \left(\frac{x_{1}-\mu_{1}}{\sigma_{1}}\right)^2} \\
  f_2(x) = \frac{1}{\sigma_{2} \sqrt{2\pi}} e^{-\frac{1}{2} \left(\frac{x_{2}-\mu_{2}}{\sigma_{2}}\right)^2} \\
  \end{aligned}
\end{equation}

\noindent where $f_1(x)$ is auto Gaussian function, and $f_2(x)$ is cross Gaussian function, and $\mu_1, \sigma_1, x_1$ and $\mu_2, \sigma_2, x_2$ are mean, standard-deviation, and x-coordinate of the intersection point of the auto and cross Gaussian distribution functions respectively.

% \begin{equation} \label{eqn:cosine_similarity}
% \begin{split}
% cosine-similarity (cos\theta) & = \frac{\textbf{X}\cdot\textbf{Y}}{\left\|\textbf{X}\right\|\left\|\textbf{Y}\right\|} \\
%  & = \frac{\sum_{i=1}^{n}X_i Y_i}{\sqrt{\sum_{i=1}^{n} X_i^2} \sqrt{\sum_{i=1}^{n} Y_i^2}}
% \end{split}
% \end{equation}

Equating and solving equations \ref{eqn:gaussian_functions} can be arranged in a condensed quadratic form: $Ax^2+Bx+C = 0$, where its coefficients are obtained as shown in equation \ref{eqn:quadratic_equation}.

\begin{equation} \label{eqn:quadratic_equation}
  \begin{split}
  A = \sigma{1}^2-\sigma{2}^2 = \nu_1-\nu_2 \\
  B = 2(\mu_1\sigma_2^2-\mu_2\sigma_1^2) = 2(\mu_1\nu_2-\mu_2\nu_1) \\ 
  C = \mu_2^2\sigma_1^2-\mu_1^2\sigma_2^2-2\sigma_1^2\sigma_2^2 \log\left(\frac{\sigma_1}{\sigma_2}\right)\\ =\nu_1\mu_2^2-\nu_2\mu_1^2-\nu_1\nu_2\log\left(\frac{\nu_1}{\nu_2}\right)
  \end{split}
\end{equation}

\noindent where, $\sigma_1, \nu_1$, and $\sigma_2, \nu_2$ are standard deviation and variance for auto and cross Gaussian distribution functions. Since, $\lambda \in \{0,1\}$, we can ignore roots lying out of this bound; otherwise, select either of the roots and evaluate the model, and consider the root with higher model accuracy. If the point of intersection lies between $\mu_{cross}$ and $\mu_{auto}$, we can proceed ahead to compute model accuracy, otherwise take an average of them as shown in equation \ref{eqn:threshold_initialization}.

\begin{equation} \label{eqn:threshold_initialization}
threshold =
\begin{cases}
  \lambda, & \text{if}\ \mu_{cross} \leq \lambda \leq \mu_{auto} \\
  \frac{\mu_{cross}+\mu_{auto}}{2}, & \text{otherwise}
\end{cases}
\end{equation}

We used different performance metrics, such as precision, recall, f1-scores, and accuracy for model evaluation. Their computation requires the count of true/false positives and negatives at given threshold $\lambda$, that are calculated as shown in the equation \ref{eqn:positives-negatives}.

\begin{equation} \label{eqn:positives-negatives}
    \begin{aligned}
        TP(\lambda) = \sum_{\imath=1}^{m} \sum_{\jmath=1}^{n} \mathds{1} \{(i,j) \in \mathbb{P}_{same}, \mathbb{S}_{max}(x_i, y_j) \geq \lambda \} \\
        FP(\lambda) = \sum_{\imath=1}^{m} \sum_{\jmath=1}^{n} \mathds{1} \{(i,j) \in \mathbb{P}_{diff}, \mathbb{S}_{max}(x_i, y_j) \geq \lambda \} \\
        FN(\lambda) = \sum_{\imath=1}^{m} \sum_{\jmath=1}^{n} \mathds{1} \{(i,j) \in \mathbb{P}_{same}, \mathbb{S}_{max}(x_i, y_j) \leq \lambda \} \\
        TN(\lambda) = \sum_{\imath=1}^{m} \sum_{\jmath=1}^{n} \mathds{1} \{(i,j) \in \mathbb{P}_{diff}, \mathbb{S}_{max}(x_i, y_j) \leq \lambda \}
    \end{aligned}
\end{equation}

\noindent where \\
$TP(\lambda)$ is the total no. of times when the model correctly predicts the positive class with a maximum similarity between feature vectors greater than equal to the threshold, 
$TN(\lambda)$ is the total no. of times when the model correctly predicts the negative class with a maximum similarity between feature vectors lesser than equal to the threshold, 
$FP(\lambda)$ is the total no. of times when the model incorrectly predicts the positive class with a maximum similarity between feature vectors greater than equal to the threshold,
$FN(\lambda)$ is the total no. of times when the model incorrectly predicts the negative class with a maximum similarity less than equal to the threshold, 
$\mathbb{S}_{max}(x_i, y_j)$ is the maximum cosine similarity between identities $x_i$ and $y_j$, $\mathbb{P}_{same}$ is auto-pairing, and $\mathbb{P}_{diff}$ is cross-pairing between the $i^{th}$ and $j^{th}$ identities.

Using the count of true/false positives and negatives at specified threshold $\lambda$ from equation \ref{eqn:positives-negatives}, model precision, recall, f1-score, and model overall accuracy can be computed as shown in equation \ref{eqn:model_metrics}.

\begin{equation} \label{eqn:model_metrics}
\begin{aligned}
    Precision(\lambda) = \frac{TP(\lambda)}{TP(\lambda)+FP(\lambda)} \\
    Recall(\lambda) = \frac{TP(\lambda)}{TP(\lambda)+FN(\lambda)} \\
    F1score (\lambda) = 2 * \frac{Precision(\lambda)*Recall(\lambda)}{Precision(\lambda)+Recall(\lambda)}\\
    Accuracy (\lambda) = \frac{TP+TN}{TP+TN+FP+FN}
\end{aligned}
\end{equation}

Similarly, true positive rate ($TPR$) and false positive rate ($FPR$) for plotting ROC curve is computed with the same count of true/false positives and negatives - resulting from the model evaluation at given threshold $\lambda$ as shown in equation \ref{eqn:roc_curve}.
\begin{equation} \label{eqn:roc_curve}
    \begin{aligned}
    TPR(\lambda) = \frac{\sum TP}{\sum CP} = \frac{TP(\lambda)}{TP(\lambda) + FP(\lambda) + \epsilon} \\
     FPR(\lambda) = \frac{\sum FP}{\sum CN} = \frac{FP(\lambda)}{FP(\lambda) + TN(\lambda) + \epsilon}
    \end{aligned}
\end{equation}

\noindent where, $CP$ is a cumulative sum of all positives count, $CN$ is a cumulative sum of all negatives count, $\epsilon$ is added to prevent from possible division by zero error.

If model accuracy at specified threshold $\lambda$ exceeds the targeted value, we can select it as an optimal threshold and proceed ahead with model inference; otherwise, we need to search for the optimal value with an objective to maximize f1-score as shown in equation \ref{eqn:optimum_threshold}.

\begin{equation}\label{eqn:optimum_threshold}
\lambda_{opt} =
\begin{cases}
  \lambda, & \text{if}\ f1score (\lambda) \geq \tau \\
  \lambda, & \text{if}\ f1score (\lambda) \geq f1score(\lambda_{old}) \\
  \lambda_{old}, & \text{otherwise}
\end{cases}
\end{equation}

For optimizing threshold at which the f1-score is below the targeted value, bounded minimization method is used to maximize the trade-off between $TPR$ and $FPR$ as shown in equation \ref{eqn:obj_func_roc}. In other words, using the optimization technique, we are trying to pick the best threshold value from the ROC curve for which we intend to maximize the number of true positives while suppressing the count of false positives.

\begin{equation} \label{eqn:obj_func_roc}
\begin{aligned}
\max_{\lambda} \quad & \left\lvert TPR(\lambda) - FPR(\lambda) \right\rvert \\
\textrm{s.t.} \quad & \lambda \in \{0,1\}    \\
\end{aligned}
\end{equation}

% \begin{equation}
% \begin{aligned}
% \max_{\lambda} \quad & \Bigg\{ {precision = \frac{TP(\lambda)}{TP(\lambda) + FP(\lambda)} \Bigg\} }\\
% \textrm{s.t.} \quad & \mu_{cross} \leq \lambda \leq \mu_{auto}\\
%   & \lambda \in \{0,1\}
% \end{aligned}
% \end{equation}

% \begin{equation}
% \begin{aligned}
% \max_{\lambda} \quad & \Bigg\{ {f1score = 2* \frac{precision(\lambda)*recall(\lambda)}{precision(\lambda) + recall(\lambda)} \Bigg\} }\\
% \textrm{s.t.} \quad & \mu_{cross} \leq \lambda \leq \mu_{auto}\\
%   & \lambda \in \{0,1\}
% \end{aligned}
% \end{equation}

However, in this study, we have directly taken maximization of f1-score as our objective function constrained to $\lambda \in \{0,1\}$ and bounded by $\mu_{cross}$ and $\mu_{auto}$. With an assumption that choosing a threshold value between the means of auto and cross-similarities density functions would maximize the number of true positives and minimize true negatives count, we tried both methods of bounded ($\mu_{cross} \leq \lambda \leq \mu_{auto}$) and unbounded optimization. As a result, it was found that unbound method i.e., $\lambda \in \{0,1\}$ resulted in higher model accuracy. Hence, we tweaked our objective function \ref{eqn:obj_func_f1score} by making search space flexible to $[0,1]$.

\begin{equation} \label{eqn:obj_func_f1score}
\begin{aligned}
\max_{\lambda} \quad & \Bigg\{ {f1score = 2* \frac{P(\lambda)*R(\lambda)}{P(\lambda) + R(\lambda)} \Bigg\} }\\
\textrm{s.t.} \quad & \lambda \in \{0,1\}
\end{aligned}
\end{equation}

At optimization termination, if precision or f1-score at the converged threshold ($\lambda$) is still less than that of the point of intersection of the Gaussian distribution functions, we choose the $\lambda_{old}$ as our working threshold for model inference as depicted in equation \ref{eqn:optimum_threshold}. This procedure as shown in figure \ref{fig:flow_diagram_adaptive_threshold} repeats if there is any change in the database, i.e., addition or deletion of any feature vector; otherwise, the chosen threshold will be used for inference unless any change in the database is observed. On the contrary, since searching optimal threshold is computationally expensive, we can run this algorithm once the model accuracy/f1-score falls below a certain value. However, for areas like medicine where a perfect threshold is a must, continuous running of this technique as illustrated in figure \ref{fig:flow_diagram_adaptive_threshold} would benefit significantly.

\subsubsection*{Model Inference}
\begin{figure}[h]
\centering
\includegraphics[width = 0.48\textwidth]{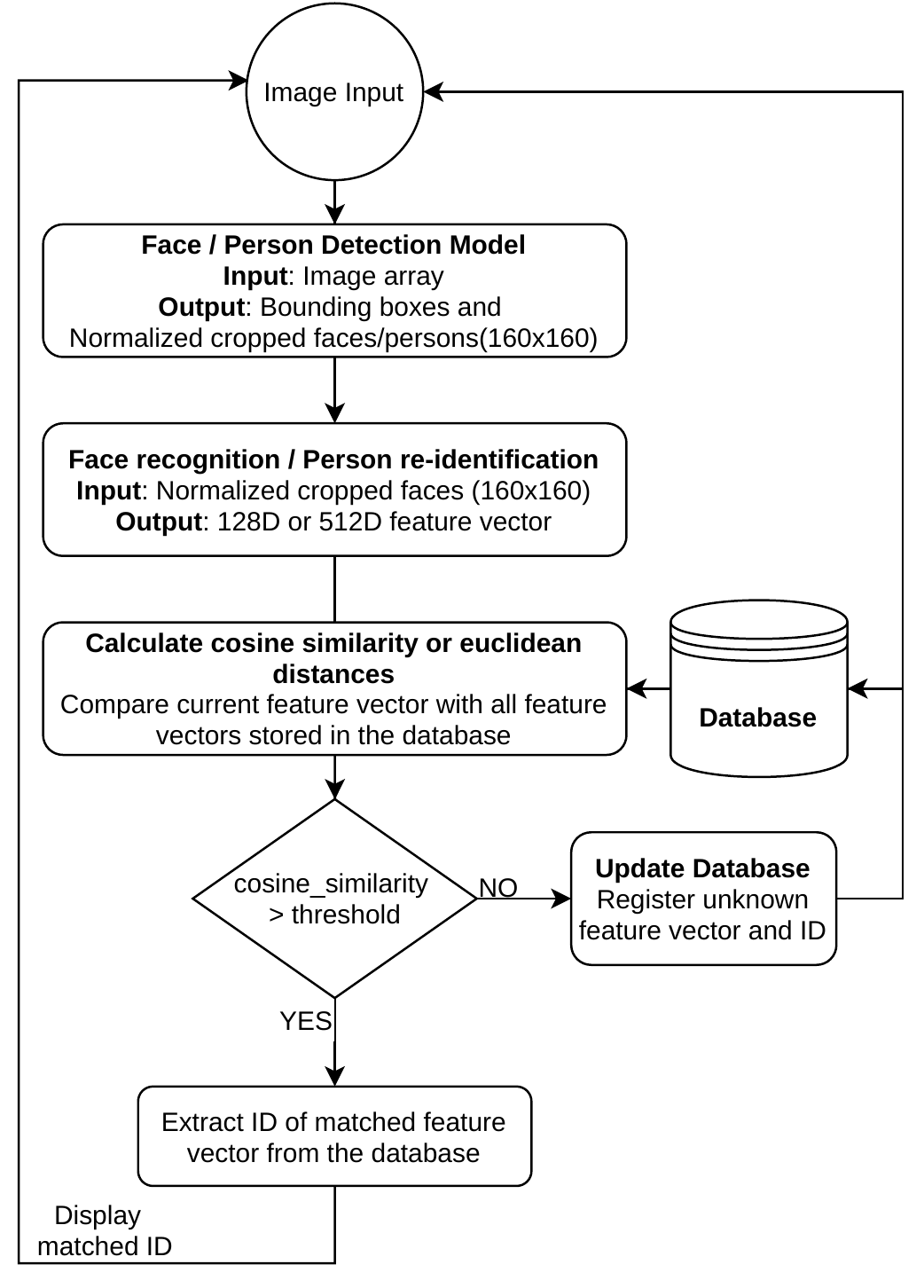}
\caption[Flow-diagram of identity recognition/re-identification]{Flow-diagram of identity recognition/re-identification}
\label{fig:face_recognition_inference}
\end{figure}

Once the optimal threshold is chosen based on an adaptive algorithm as portrayed in figure \ref{fig:flow_diagram_adaptive_threshold}, we proceed ahead with model inference. In this section, as unfolded in figure \ref{fig:face_recognition_inference}, the feature vector of an image either from webcam or directly from the image itself is generated using pre-trained recognition or re-identification model. Here, for face recognition, facenet \cite{Schroff} and dlib \cite{King2009} models are taken into account, and joint discriminative and generative learning model for person re-identification. For a given query image, we first detect face or person alignment using CNN models. Once a face is detected, we feed it into the pre-trained model to extract its 128D or 512D facial feature vector. Consequently, we perform dot product of this feature vector with an entire pool of vectors stored in the database to calculate the similarities between each of them. Upon comparison of the similarity metric for a query vector, if it is greater than the chosen decision threshold, it is assumed that the identity has been identified or recognized, and model inference continues. Conversely, if the similarity of the query identity does not match with any of the gallery vectors, it is recorded as a new identity in the database.

\section{Experiments}
This method has been tested on labeled faces in the wild (LFW) \cite{Huang}, that is a de facto face verification dataset. Since most of the identities in the LFW dataset comprise at most two images per identity, full effectiveness and robustness of an adaptive threshold algorithm is not observed. A separate dataset of the top 100 highly-paid athletes referred to Forbes's 2018 list \cite{forbes_athletes} is prepared with more than 20 images per identity with an online automated dataset generator \cite{online_dataset_generator}. Since the same set-of-identities is used for galley and query embeddings, only identities with two or more than two images are taken into account. On the other hand, since the whole model accuracy depends on the number of images covering an entire $180^\circ$ facial orientation of a person, there needs to be a flexibility in adjusting it. If headpose is incorporated, we can choose face oriented only at a particular direction so that 3-4 images would cover entire possible regions of detection. For instance, the most accurate CNN-based face detector can detect a face rotated up to $90^\circ$ in either direction w.r.t a perpendicular line protruded out of the face. The other important point to be noted is that since the same dataset, i.e., LFW and Athletes, is used for model evaluation, to simulate a real-time dynamic nature of the database or gallery size, the number of identities is increased one-by-one starting with two at the beginning. That is to say that during the start of the algorithm, two identities are accounted to compute a threshold; then substantially all the available identities are added to the galley one at a time; simultaneously computing the appropriate threshold with the momentary database size. Hence, using the proposed algorithm, the threshold learns to adapt to a changing gallery/database identities' size.

\begin{figure}[h]
\centering
\includegraphics[width = 0.48\textwidth]{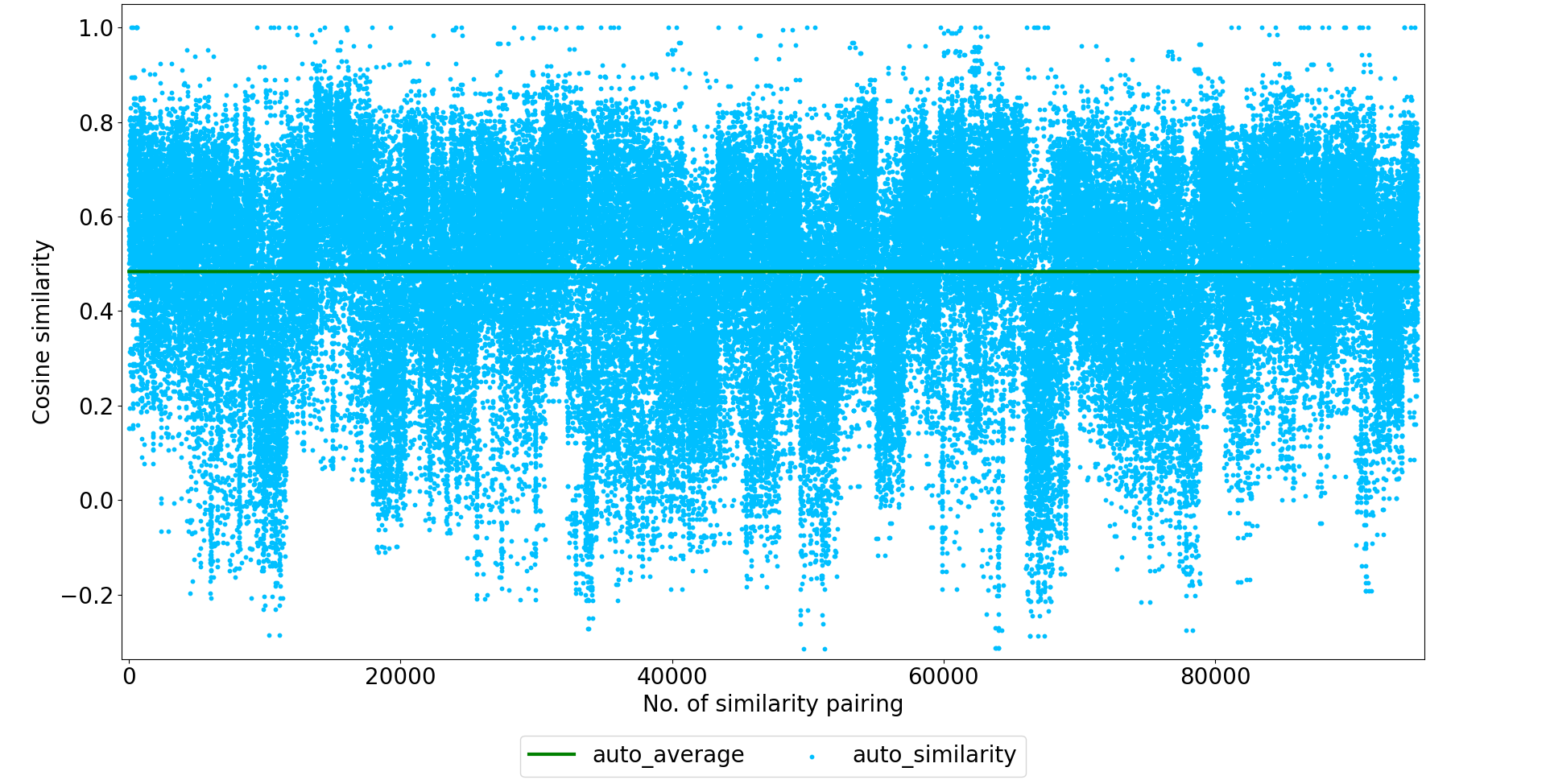}
\caption[Auto-similarity distribution among faces of same identities]{Auto-similarity distribution among faces of same identities}
\label{fig:auto_similarity_distribution}
\end{figure}

From figure \ref{fig:auto_similarity_distribution}, cosine similarities between the same identities (auto) is distributed in such a way that maximal distribution lies within the range of $[0.2-0.8]$.

\begin{figure}[h]
\centering
\includegraphics[width = 0.48\textwidth]{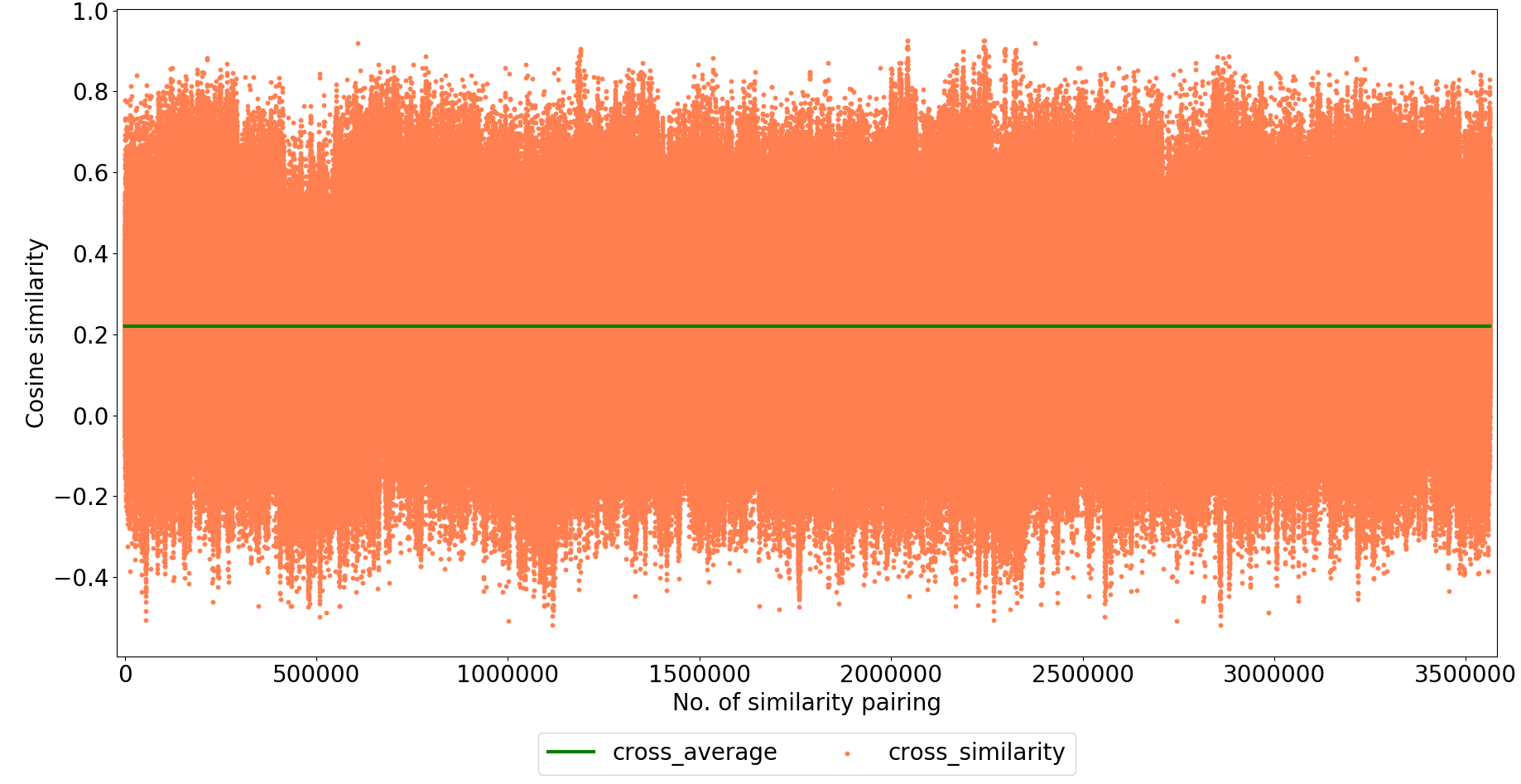}
\caption[Cross-similarity distribution among faces of different identities]{Cross-similarity distribution among faces of different identities}
\label{fig:cross_similarity_distribution}
\end{figure}

\begin{figure}[h]
\centering
\includegraphics[width = 0.48\textwidth]{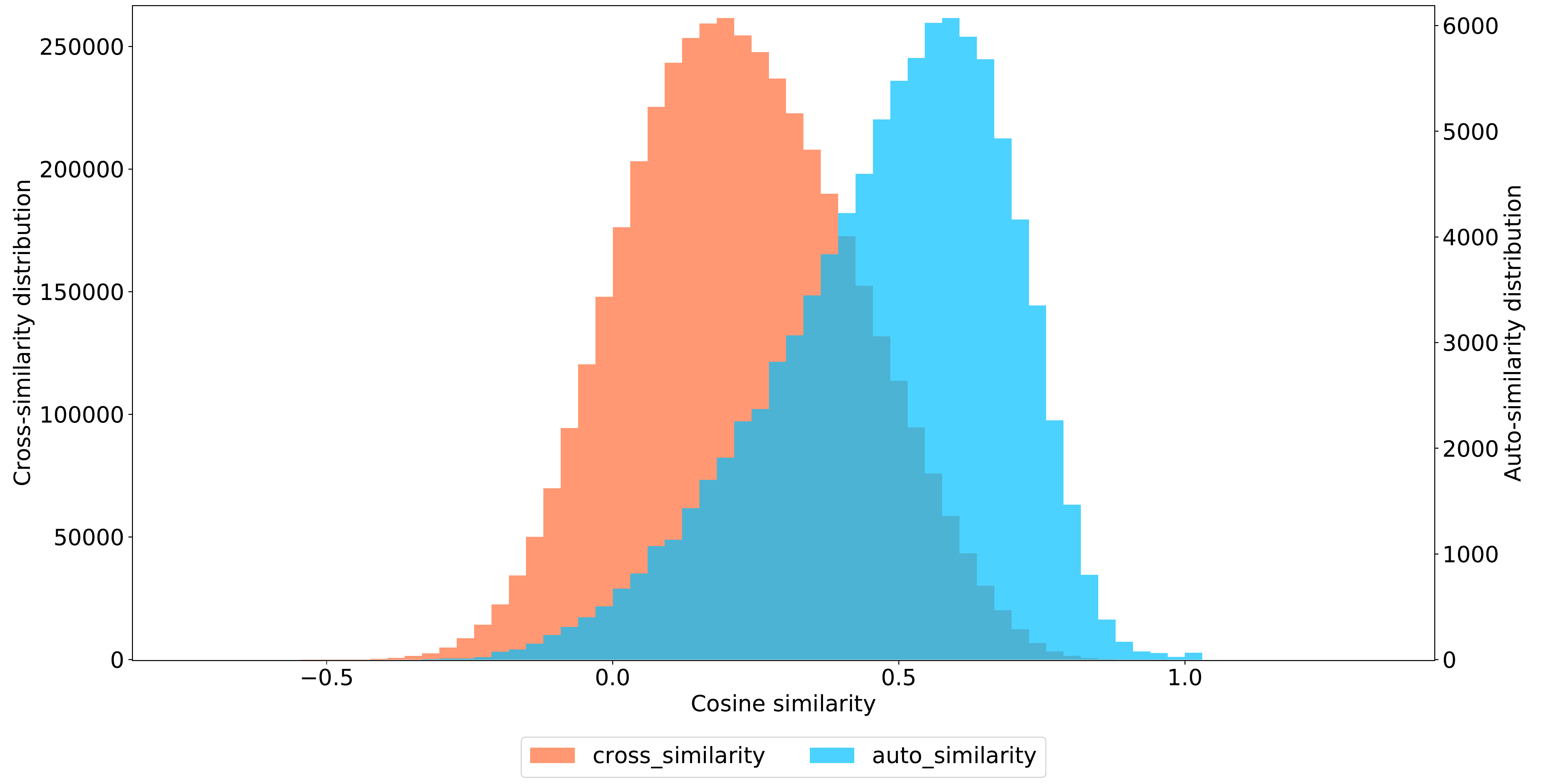}
\caption[Probability of occurrences of similarities among same (auto) and different identities (cross)]{Probability of occurrences of similarities among same (auto) and different identities (cross)}
\label{fig:auto_cross_histogram}
\end{figure}

From figure \ref{fig:cross_similarity_distribution}, it is observed that mostly similarity among different identities is distributed between the range $[-0.4, 0.8]$. In reference to the distribution pattern of the auto/cross similarities observed in figures [\ref{fig:auto_similarity_distribution}, \ref{fig:cross_similarity_distribution}], and histogram \ref{fig:auto_cross_histogram}, if a single threshold ought to be chosen would be in between their means i.e., $[0.2-0.6]$. If we chose the lower bound of a choice list, the threshold would perform better at identifying similar identities, whereas bad at differentiating dissimilar ones, and vice versa for the upper bound. Therefore, in this project, we have used an online optimization-based threshold technique that can adapt to the changes in the database size and their probability density functions.

\begin{figure}[h]
\centering
\includegraphics[width = 0.48\textwidth]{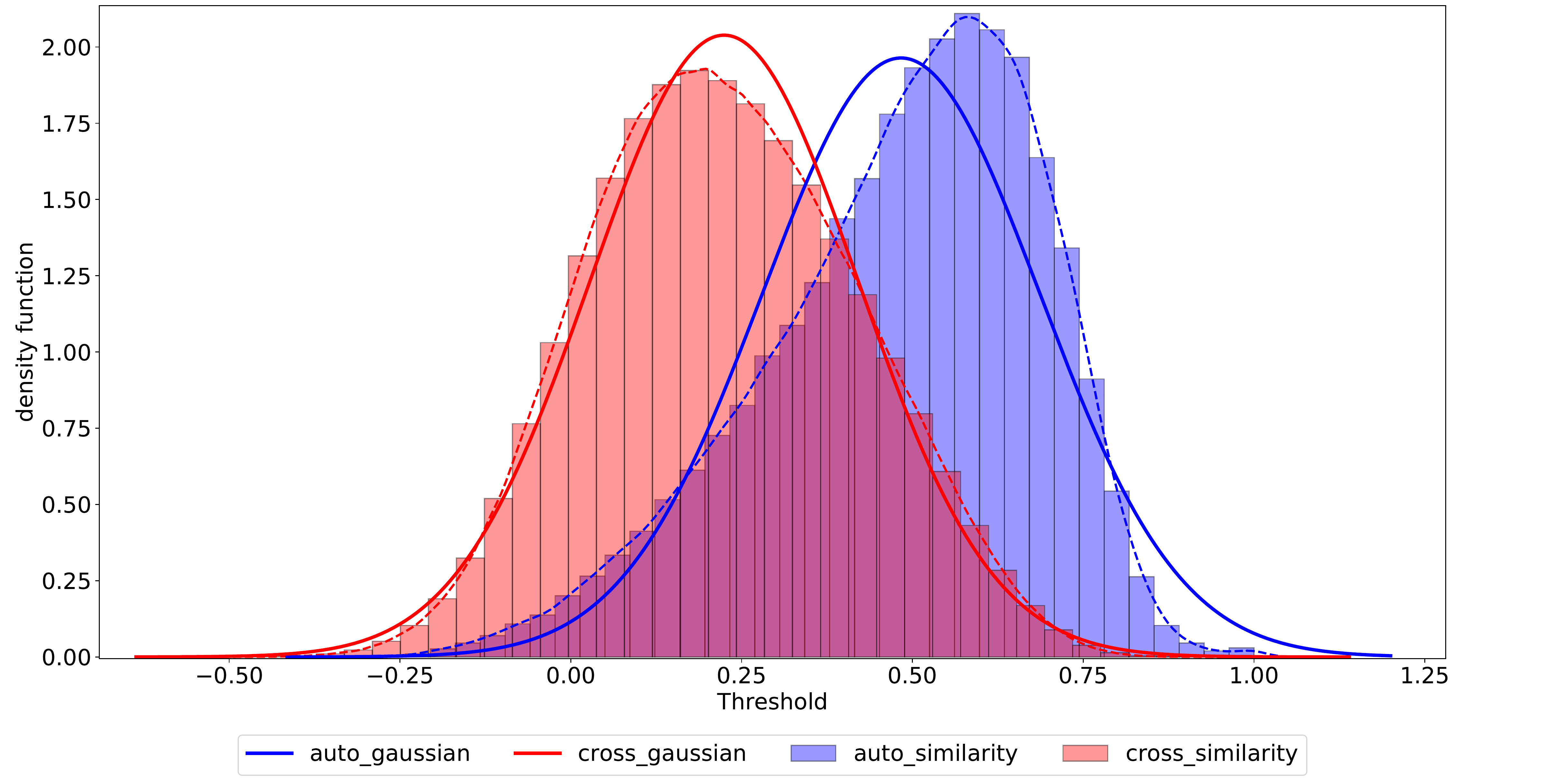}
\caption[Estimated actual and Gaussian distribution functions for standardized auto and cross similarities functions]{Estimated actual and Gaussian distribution functions for standardized auto and cross similarities functions}
\label{fig:auto_cross_gaussian_functions}
\end{figure}

Figure \ref{fig:auto_cross_histogram} shows probability density of similarity function for same and different identities for an Athletes dataset. From the observed histogram, a normal distribution function is estimated using a mean and standard deviation of the actual auto and cross similarity distributions. Figure \ref{fig:auto_cross_gaussian_functions} shows that the point of intersection of the actual auto and cross similarity distribution functions separates the curves in a most unbiased way. Higher model accuracy is assumed to be achieved if it is taken as a threshold. There is no such defined function for those actual probability distribution functions. However, their estimated Gaussian distribution function mimics the distribution to a larger extent, though there is a slight shift in the point-of-intersection. The intersection point of the estimated normal distribution functions is computed by solving a quadratic equation with coefficients shown in equation \ref{eqn:quadratic_equation}. This point can be taken as an initial value for an adaptive threshold and perform its conditional inspection. If model accuracy (here f1-score) is greater than the targeted value (here $\geq 80\%$), we take it as an optimal threshold, whereas if it is not, then we proceed ahead with an optimum search via bounded optimization method as shown in the objective function \ref{eqn:obj_func_f1score}. Upon the termination of an optimal search, if f1-score at that point ($\lambda$) is greater than that of the point of intersection of the estimated Gaussian functions, it is taken as an optimal threshold; otherwise, the intersection point ($\lambda_{old}$) is taken as an optimal threshold.

\begin{figure}[h]
\centering
\includegraphics[width = 0.48\textwidth]{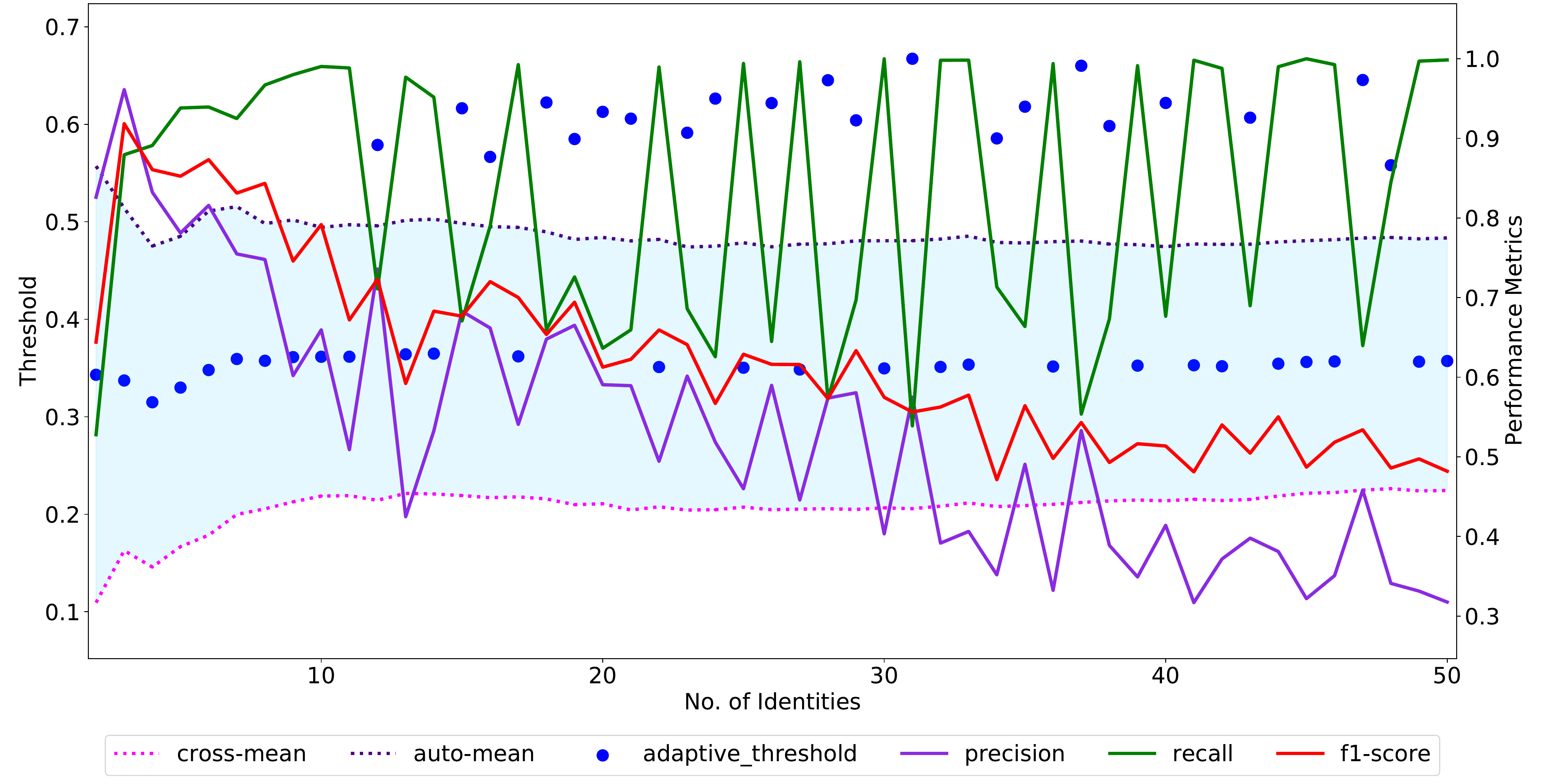}
\caption[Model performance at different threshold (adaptive)]{Model performance at different threshold (adaptive)}
\label{fig:adaptive_metrics_comparitive}
\end{figure}

\begin{figure}[h]
\centering
\includegraphics[width =0.48\textwidth]{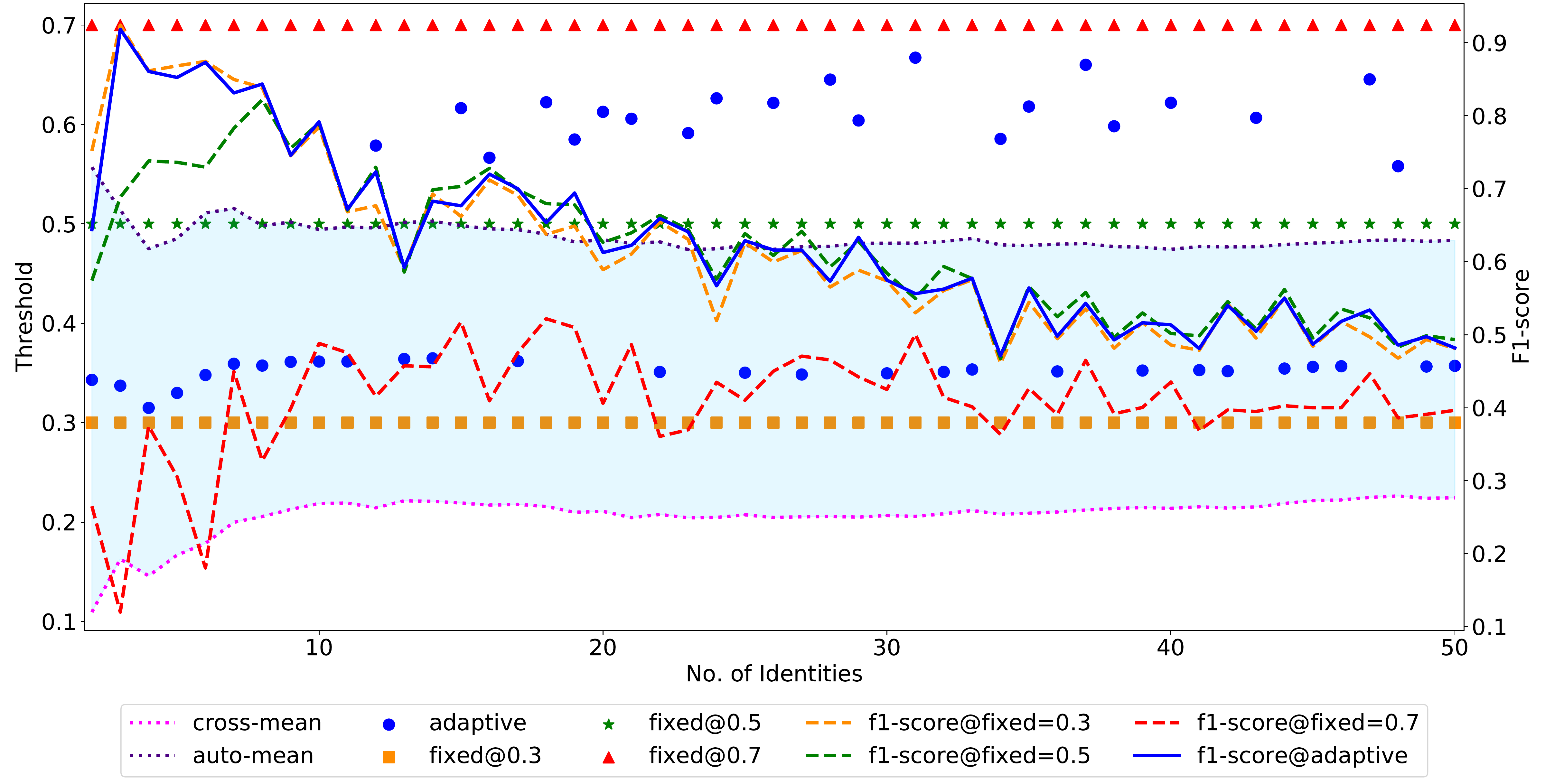}
\caption[Comparative study of fixed threshold @\{0.3, 0.5, 0.7\} with adaptive threshold]{Comparative study of fixed threshold @\{0.3, 0.5, 0.7\} with adaptive threshold}
\label{fig:comparative_threshold_vs_f1score}
\end{figure}

The model performance at the chosen optimum threshold is compared with its performance at various fixed threshold values that are usually taken via hit-and-trial method. In this study, the pre-specified fixed threshold points are chosen from auto and cross similarity distribution functions, i.e., point at maximum density or occurrences. From the auto and cross probability density function referred to figure \ref{fig:auto_cross_histogram}, $\mu_{auto}$ and $\mu_{cross}$ for the Athletes dataset is observed to be 0.3 and 0.6, respectively. From figure \ref{fig:comparative_threshold_vs_f1score}, it is observed that model accuracy at adaptive threshold overshadows the performance at fixed threshold ($=0.5$) (which is the as-usual chosen threshold in the context of conventional classification problem) and $0.7$ - at which model precision is highest among all. However, looking at f1-score, model performance at adaptive threshold is more likely comparable to that of the fixed threshold ($=0.3$). It might be because of its higher recall value; in contrast, looking at the comparative precision curve, it is worst among all the chosen threshold, including adaptive one. Hence, therefore, inspecting precision, recall, and f1-score separately, adaptive threshold outperforms all the chosen fixed thresholds believed that they are the best and conventionally taken as-usual thresholds. From the figure \ref{fig:comparative_threshold_vs_f1score}, one of the noticeable things of an adaptive threshold is that once the number of identities in the gallery increases, resulting in a drastic drop in the model accuracy, there is an abrupt adaptation in the threshold to keep the model constraints within check. In this particular test case, model accuracy at the fixed threshold ($=0.3$) is quite similar to that of an adaptive threshold, however, for a large number of identities in the database, its adaptive nature will start outperforming rest of all the fixed values. Hence, with a larger dataset or in case of continuous running of the model, it is evident that gallery size will increase indefinitely; consequently, the adaptive threshold will adjust for better performance while the best-fixed threshold deteriorates relentlessly over time-span.

\begin{figure}[h]
\centering
\includegraphics[width = 0.48\textwidth]{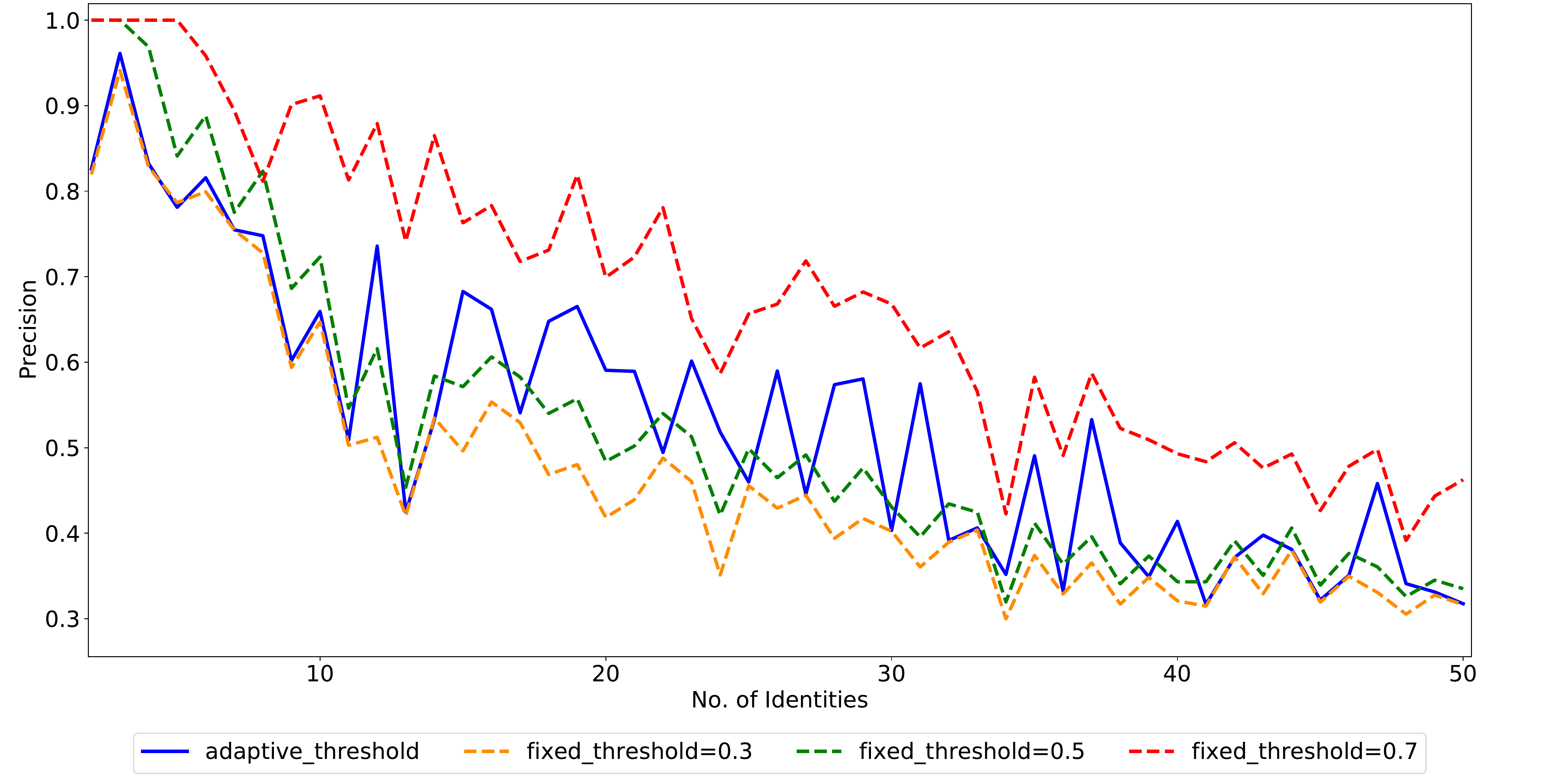}
\caption[Precision curve at fixed threshold @\{0.3, 0.5, 0.7\} and adaptive threshold]{Precision curve at fixed threshold @\{0.3, 0.5, 0.7\} and adaptive threshold}
\label{fig:comparative_study_precision}
\end{figure}

\begin{figure}[h]
\centering
\includegraphics[width = 0.48\textwidth]{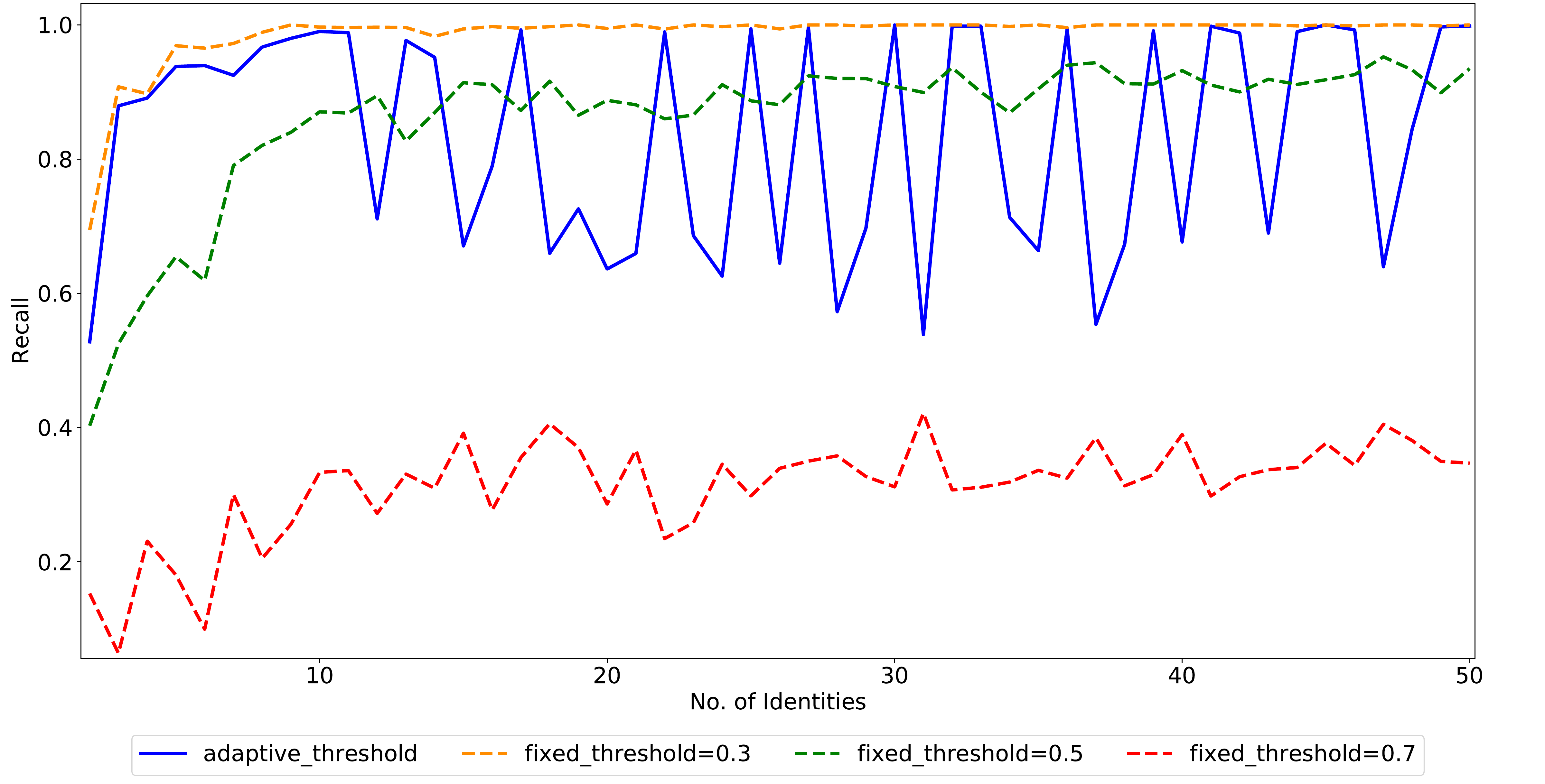}
\caption[Recall curve at fixed threshold @\{0.3, 0.5, 0.7\} and adaptive threshold]{Recall curve at fixed threshold @\{0.3, 0.5, 0.7\} and adaptive threshold}
\label{fig:comparative_study_recall}
\end{figure}

From figure \ref{fig:comparative_study_precision}, it is observed that the model precision is highest at fixed threshold ($=0.7$) followed by fixed threshold ($=0.5$), but from figure \ref{fig:comparative_study_recall}, recall at those values are worst. Hence, looking at the harmonic mean of precision and recall at those fixed points can be discarded. Among all of those chosen points, model performance is good at the adaptive threshold and fixed threshold ($=0.3$). Since this experiment has been done only on a small dataset, its performance at the lowest fixed threshold seems comparable to that of adaptive. Nevertheless, there is a higher probability of getting worse for a larger dataset because of its worsening precision with drastically increasing database size.

\begin{figure}[h]
\centering
\includegraphics[width = 0.48\textwidth]{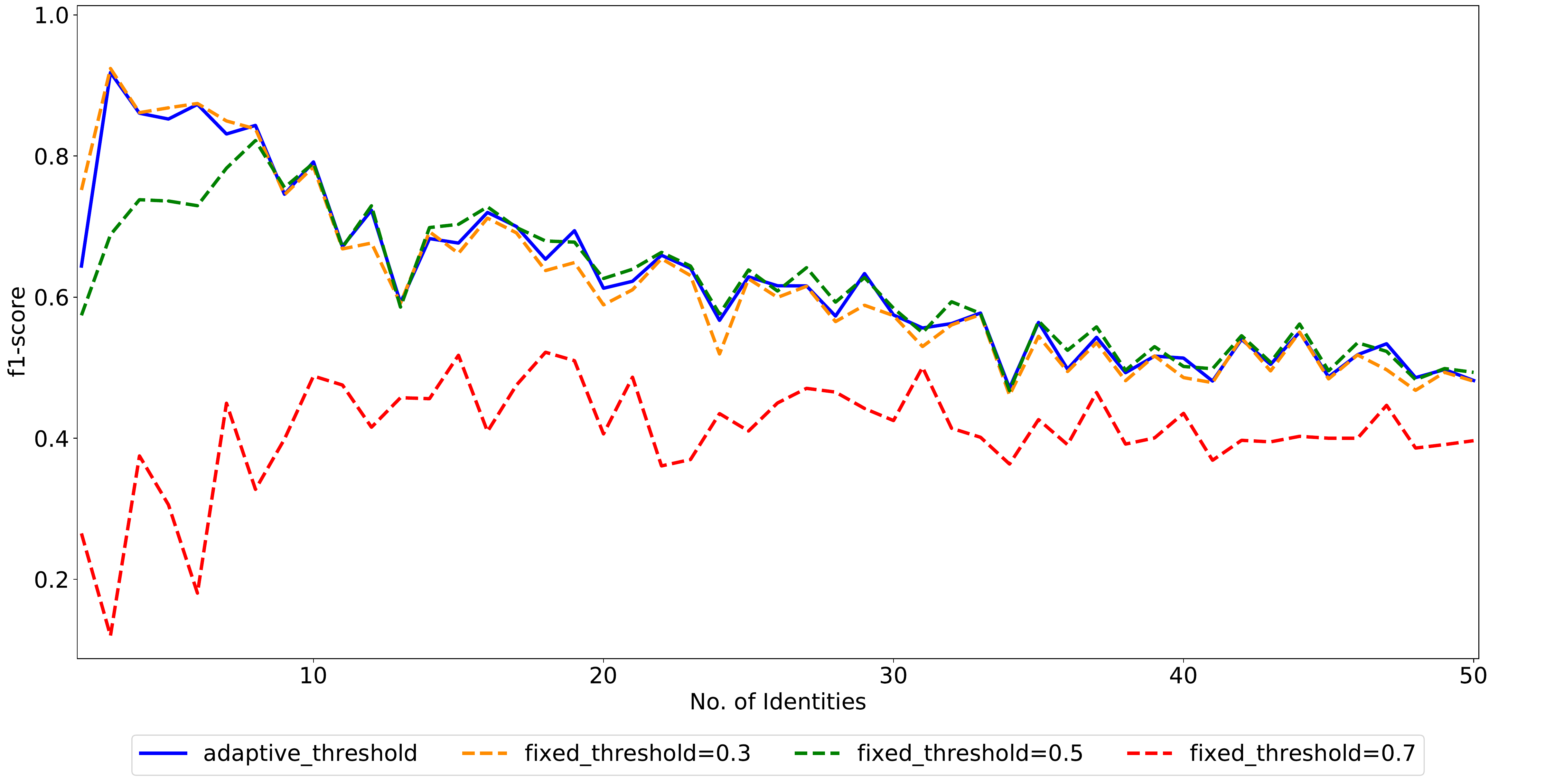}
\caption[F1-score curve at fixed threshold @\{0.3, 0.5, 0.7\} and adaptive threshold]{F1-score curve at fixed threshold @\{0.3, 0.5, 0.7\} and adaptive threshold}
\label{fig:comparative_study_f1score}
\end{figure}

\begin{figure}[h]
\centering
\includegraphics[width = 0.48\textwidth]{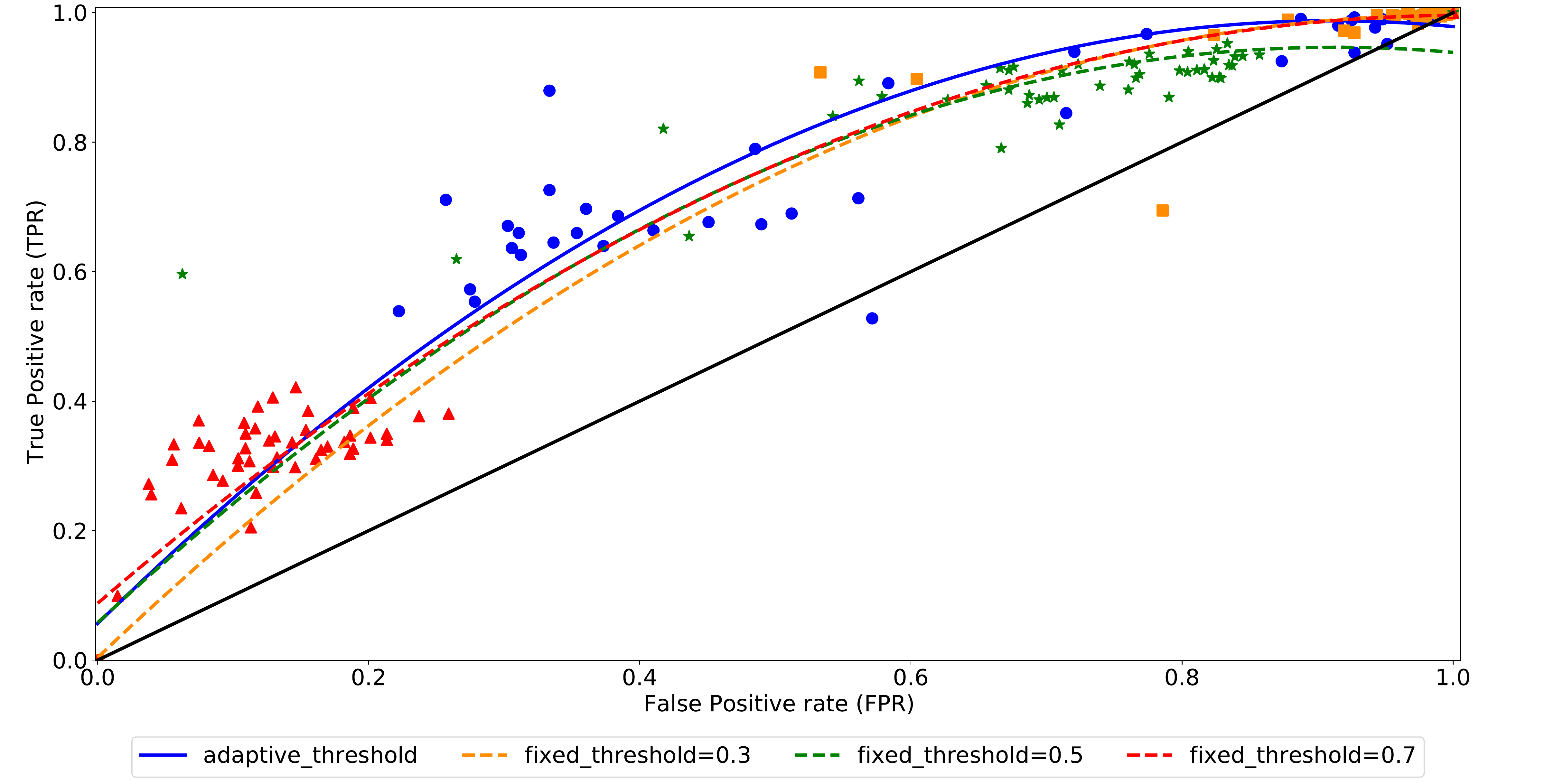}
\caption[ROC curve at fixed threshold @\{0.3, 0.5, 0.7\} and adaptive threshold]{ROC curve at fixed threshold @\{0.3, 0.5, 0.7\} and adaptive threshold}
\label{fig:roc_curve}
\end{figure}

From ROC curve \ref{eqn:roc_curve}, it is observed that the ratio of true positives to false positives is higher with an adaptive threshold as compared to any of the fixed thresholds $\{0.3. 0.5, 0.7\}$. Looking at the curve, if we ought to choose any one of the fixed thresholds, then it would be $0.5$ as its area under the curve is highest among all the fixed ones. Thus, the ROC curve's comparative plot validates that the adaptive method is a good choice for choosing a threshold for the chosen dataset. 

% Please add the following required packages to your document preamble:
% \usepackage{booktabs}
\begin{table}[]
\centering
\caption{Comparative model performance at different thresholds}
\label{tab:model_performance}
\begin{tabular}{@{}cccc@{}}
\toprule
\begin{tabular}[c]{@{}c@{}}Threshold\\ Type\end{tabular} &
  \begin{tabular}[c]{@{}c@{}}AUC \\ (unit sq.)\end{tabular} &
  \begin{tabular}[c]{@{}c@{}}Model\\ Accuracy\\ (\%)\end{tabular} &
  \begin{tabular}[c]{@{}c@{}}F1score $\geq$ 0.8\\ (\%)\end{tabular} \\ \midrule
Adaptive  & 0.72 & 88.34 & 20.32 \\
fixed@0.3 & 0.62 & 66.89 & 19.54  \\
fixed@0.5 & 0.68 & 78.22 & 5.31  \\
fixed@0.7 & 0.64 & 60.53 & 0.0   \\ \bottomrule
\end{tabular}
\end{table}

Table \ref{tab:model_performance} shows that the model performs well with an adaptive threshold compared to any of the fixed thresholds. Though f1-score is $\geq 80\%$ for most of the time for a fixed threshold at ($=0.3$), its precision is horrible. The higher f1-score might be because of its higher recall value. On the contrary, a fixed threshold of ($=0.7$) does have higher precision than any of the other bands, as vivid in figure \ref{fig:comparative_study_precision}, but due to its bad recall, the final f1-score is lowest among all. If we look solely at the model accuracy, as shown in equation \ref{eqn:model_metrics}, a model with an adaptive threshold outperforms all the other fixed threshold values. Similarly, if we look at the area under the ROC-curve, the adaptive threshold occupies maximum area that further validates model's better performance with the adaptive threshold.

\section{Conclusion}
Most of the research in machine learning focuses on improving the existing model architectures and modifying the optimization algorithms and loss functions. However, there is hardly any research done in making the decision threshold adaptive. Once the model outputs confidence scores, they need to be sieved with a decision threshold to categorize them to their respective classes. If the threshold is not chosen wisely, the entire model performance will go in vain. However, in most cases, this significant deciding value is taken via hit-and-trial method; or randomly some fixed value just by analogy (mostly 0.5 by default). Mostly for examples like classifying different objects - where each object's features are vividly distinct, it might work, but not for all cases.

Nonetheless, for identification tasks like in face-recognition and person re-identification - where features between different identities are more or less identical and inseparable, the conventional method of choosing a fixed threshold might not work. To counteract the idea of selecting an optimum threshold based on ROC-curve, it might fail for temporally increasing identity-database size. For a deep learning model that encounter classification/identification of the objects with a higher degree of features-similarity, a threshold that could adapt to the varying database size would play a crucial role. Therefore, an online optimization-based statistical feature learning technique is proposed to formulate an adaptive threshold in this project. Whenever there is any update on the identity-database size, a new decision threshold will be calculated based on the estimated Gaussian distribution functions of the auto/cross-similarity distributions and bound-constrained optimization search method. The proposed algorithm was tested on Labeled Faces in the Wild (LFW), a de-facto face verification dataset, and a self-prepared dataset of top 100 highly-paid athletes published by Forbes magazine (2018). The reason behind preparing own dataset is that the algorithm requires a large number of samples per identity for probing its robustness over widely distributed features. The method achieved $32.07\%$, $12.94\%$, and $45.94\%$ higher accuracy compared to the fixed thresholds at $0.3$, $0.5$, and $0.7$, respectively. Looking at model accuracy, with adaptive threshold f1-score was observed to $20.32\%$ times $ \geq 80\%$ w.r.t the total number of samples; whereas $19.54\%$, $5.31\%$, and $0.00\%$ for fixed thresholds at 0$.3$, $0.5$, and $0.7$ respectively. This project has been tested on a concise dataset, so in the future, researchers can explore more on such techniques that can make decision threshold adapt to varying identities-database size and indistinguishable features.

% from the probability density functions of the similarity between same (here denoted as auto-similarity) and different (cross-similarity) identities

\section*{Acknowledgement}
The author declares that the research was conducted in the absence of grants or financial support from any institution or organization. However, deHumlaTech AI Research team would like to thank Mr. Bibek Raj Shrestha and Mr. Baibhav Raj Shrestha, for providing the necessary hardware components like RAM-16GB and SSD-500GB to encounter its higher computational requirements.

% \\\\\\\\\\\\\\\\\\\\\\\\\\\\\\
% \section{References}
% {\small
% \bibliographystyle{ieee_fullname}
% \bibliography{references}
% }

{\small
\bibliography{references.bib}{}
\bibliographystyle{IEEEtran}
}

\end{document}